%% file: template.tex
\documentclass{article}

\usepackage{arxiv}

\usepackage[utf8]{inputenc} 
\usepackage[T1]{fontenc}    
\usepackage{hyperref}       
\usepackage{url}            
\usepackage{booktabs}       
\usepackage{amsfonts}       
\usepackage{nicefrac}       
\usepackage{microtype}      
\usepackage{lipsum}
\usepackage{graphicx}
\usepackage{multirow}
\usepackage{array}
\graphicspath{ {./images/} }

\usepackage{times}  
\usepackage{helvet}  
\usepackage{courier}  
\usepackage{graphicx} 
\usepackage{caption} 
\usepackage{algorithm}
\usepackage{algorithmic}
\usepackage[textwidth=1.5cm]{todonotes}

\usepackage{newfloat}
\usepackage{listings}
\DeclareCaptionStyle{ruled}{labelfont=normalfont,labelsep=colon,strut=off} 
\floatstyle{ruled}
\newfloat{listing}{tb}{lst}{}
\floatname{listing}{Listing}

\title{Advancing Single and Multi-task Text Classification through Large Language Model Fine-tuning}

\author{
  Hang Zhao\thanks{These authors contribute equally to this paper.} \\
  Meta Platforms, Inc.\\
  Menlo Park, CA 94025 \\
  \texttt{hangz@meta.com} \\
   \And
  Qile P. Chen\footnotemark[1]\\
  Meta Platforms, Inc.\\
  Menlo Park, CA 94025 \\
  \texttt{pqchen@meta.com} \\
   \And
  Yijing Barry Zhang\footnotemark[1] \\
  Anthropic PBC\footnotemark[2]\thanks{Work done at Meta Platforms, Inc. Currently at Anthropic PBC.}\\
  San Francisco, CA 94104 \\
  \texttt{Barryz@anthropic.com} \\
   \And
  Gang Yang\footnotemark[1] \\
  Meta Platforms, Inc.\\
  Menlo Park, CA 94025 \\
  \texttt{gangy@meta.com} \\
}

\begin{document}

\maketitle
\input{sections/0_abstract}
\input{sections/1.0_introduction.tex}

\input{sections/2.0_related_work.tex}

\input{sections/3.0_Models.tex}
\input{sections/3.1_Experiments.tex}

\input{sections/4.0_general_discussion.tex}

\input{sections/5.0_conclusion.tex}

\input{references.tex}
\input{sections/6.0_appendix.tex}
\end{document}

%% file: sections/0_abstract.tex
\section{Abstract}

Both encoder-only models (e.g., BERT, RoBERTa) and large language models (LLMs, e.g., Llama3) have been widely used for text classification tasks. However, there is a lack of systematic studies comparing the performance of encoder-based models and LLMs in text classification, particularly when fine-tuning is involved. This study employed a diverse range of models and methods, varying in size and architecture, and including both fine-tuned and pre-trained approaches. We first assessed the performances of these LLMs on the 20 Newsgroups (20NG) and MASSIVE datasets, comparing them to encoder-only RoBERTa models. Additionally, we explored the multi-task capabilities of both model types by combining multiple classification tasks, including intent detection and slot-filling, into a single model using data from both datasets. Our results indicate that fully fine-tuned Llama3-70B models outperform RoBERTa-large and other decoder LLMs across various classification tasks and datasets. Moreover, the consolidated multi-task fine-tuned LLMs matched the performance of dual-model setups in both tasks across both datasets. Overall, our study provides a comprehensive benchmark of encoder-only and LLM models on text classification tasks and demonstrates a method to combine two or more fully fine-tuned decoder LLMs for reduced latency and equivalent performance.



%% file: sections/1.0_introduction.tex
\section{Introduction}
Recent advancements in Large Language Models (LLMs) have yielded state-of-the-art results across a broad spectrum of Natural Language Processing (NLP) tasks, including text classification, summarization, and question answering~\cite{radford2019language,ouyang2022training}. The concurrent emergence of proprietary models (e.g., GPT-4~\cite{achiam2023gpt, openai_gpt4_research}, Claude~\cite{anthropic2024claude}, Gemini~\cite{reid2024gemini}) and openly-available alternatives (e.g., Llama2~\cite{touvron2023llama}, Llama3~\cite{MetaAI2024}, Mistral~\cite{jiang2023mistral, mistral_nemo_2024}, Nemotron~\cite{parmar2024nemotron, adler2024nemotron}) has catalyzed exploration of innovative downstream applications, such as autonomous agents~\cite{wang_survey_2023}, underscoring the vast potential of LLMs in transforming the NLP landscape.

Text classification, a fundamental task in NLP, involves categorizing text into predefined categories. This task is critical and widely used in numerous NLP applications, encompassing various classic NLP tasks, but not limited to: sentiment analysis, topic classification, spam detection, intention classification, textual entailment, document classification, fake news detection, hate speech detection, relation extraction, paraphrase identification, emotion detection, sarcasm detection, language identification, and named entity recognition (NER). Furthermore, in more complex AI agents, text classification serves as a base module, widely employed to ensure the appropriate tool or agent is triggered, with examples including intent detection, tool selection, and agent routing.

Both encoder-only models (e.g., BERT, RoBERTa)~\cite{devlin2018bert, sun2019fine, yang2019xlnet, liu2019roberta, soyalp2021improving} and encoder-decoder models (e.g., T5)~\cite{liu2021enct5, mucke2023fine, zhuang2023rankt5} are extensively utilized for text classification tasks. For these models, fine-tuning is essential to attain high performance. More recently, leveraging LLMs for classification tasks has garnered attention in research fields and industries. LLMs, such as Llama2, GPT-3.5, ChatGPT, GPT4, and Claude, have been adopted as classifiers~\cite{yu2023open, sun2023text, bucher2024fine} for these tasks. Compared to encoder-only and encoder-decoder models, LLMs possess the unique advantage of demonstrating stronger performance without fine-tuning, where a well-designed prompt can effectively guide LLMs in classification tasks~\cite{yu2023open}. Nevertheless, LLM fine-tuning is also considered a common strategy to direct the model to achieve expected outcomes. While fine-tuning generally yields superior performance in classification tasks, it requires significantly more computational resources and a larger amount of data during training compared to prompt engineering.

Beyond single-task text classification, numerous studies ~\cite{chen2019bert,korpusik2019comparison,qin2019stack,zhang2019joint,castellucci2019multi,liu2019cm,pentyala2019multi,krone2020learning,ni2020natural,tang2020end,ren2020intention,wang2020sasgbc,tu2023joint} have focused on using encoder models for multi-task classification, such as joint intent detection and slot filling tasks, achieving higher accuracy likely due to shared embedding representations. To accommodate multi-task prediction capabilities, one must customize the model architecture to add different prediction heads for encoder models. LLMs, conversely, are inherently multi-task learners~\cite{radford2019language} and possess the flexibility to define output structures for multi-task predictions without altering the model architecture.

Previous work~\cite{bucher2024fine} demonstrated that the fine-tuned encoder models such as RoBERTa-large and DeBERTa-V3 significantly outperformed LLMs without fine-tuning (e.g., ChatGPT with GPT-3.5/GPT-4 and Claude Opus) on the text classification tasks. There is also a study ~\cite{sun2023text} indicating that few-shot LLMs achieved better performance than zero-shot LLMs in text classification. However, there are few systematic studies on the text classification performance of LLMs and their comparison with encoder-only models. Moreover, there is a lack of research on performance benchmarking and comparison of multi-task classification tasks for encoder and LLM models. Our study filled these gaps by systematically exploring the key levers in determining the single- and multi-task text classification performance of LLMs, and comparing their performance with encoder-only models on openly-available classification datasets.

In this research, we comprehensively explore open, closed, small, and large decoder LLMs on single- and multi-task text classification, comparing them with pre-trained instruction-tuned decoder LLMs and traditional BERT-like models. We also investigate various setups using different LLMs, with and without fine-tuning, model sizes, model versions, prompt setups, and zero-shot and few-shot approaches. We provided insights into understanding key levers to improve LLMs' text classification performance. The main contributions of this study are two-fold: (1) the fine-tuned Llama3-70B with five-shot prompting has achieved the best performance on both single- and multi-task text classification datasets among all the LLMs, outperforming encoder models such as RoBERTa-large~\cite{liu2019roberta}; (2) the multi-task classification model with consistent input has achieved performance on par with dual-model systems while reducing the number of model calls into one.

%% file: sections/2.0_related_work.tex
\section{Related Work}

\subsection{Pre-trained LLM for Text Classification}
The application of LLMs to text classification tasks has evolved significantly. Early works of BERT~\cite{devlin2018bert} and RoBERTa~\cite{liu2019roberta} demonstrated the efficacy of pre-trained encoder models. Subsequently, decoder-only architectures, such as GPT-3~\cite{brown2020language} exhibited promising capabilities in zero-shot and few-shot paradigms. In more recent works, LLMs with few-shot learning have been shown to outperform classical ML methods and small language models like DeBERTa  examples~\cite{bucher2024fine}. Further improvements, either in prompting techniques like chain-of-thought reasoning and CARP~\cite{sun2023text}, or through automated few-shot examples~\cite{aly2023automated} can further improve performances on a variety of text classification tasks.

\subsection{Fine-tuning LLMs for Text Classification}
While LLMs are known for their ability to generalize, given text classification's often supervised nature, there is a rich tradition in utilizing fine-tuning for text classifications. Early work like ULMFiT~\cite{howard2018universal} showed as few as 100 examples can drastically improve accuracy. In recent works, fine-tuned small language models generally outperform larger LLMs with few-shot learning~\cite{edwards2024language}. Fine-tuned openly-available models have also achieved parity with closed-source models without fine-tuning~\cite{yu2023open}~. Interestingly, task-specific small models frequently demonstrate comparable or superior performances to their larger counterparts.~\cite{bucher2024fine, yu2023open}~This is one of the results we try to understand in our work.

\subsection{Multi-task Learning for LLMs}
Multi-task learning (MTL) has emerged as a powerful paradigm in deep learning. The ability to leverage shared knowledge across multiple tasks offers clear advantages such as data efficiency and reduced over-fitting~\cite{crawshaw2020multi}. Early explorations of MTL training on BERT models already showed the efficacy of the approach~\cite{chen2019bert}, but LLMs brought infinite possibilities when they demonstrated strong unsupervised multi-task learning abilities~\cite{radford2019language}. Since then, various approaches to MTL have been widely explored, including the multi-task learning stage between pretraining and fine-tuning stages, termed pre-finetuning ~\cite{aghajanyan2021muppet}, dynamic routing of low-rank adaptations~\cite{huang2023lorahub}, and mixture-of-experts~\cite{shen2023mixture, feng2024mixture, yang2024moral, luo2024moelora, dou2023loramoe, gou2023mixture, liu2023moelora, gao2024higher, jiang2024mixtral, wu2024parameter}. 

More recent work established benchmarks for multi-task inference ~\cite{son2024multi} and found surprising results that LLMs had improved performances when given multiple task instructions at once. While the previous works~\cite{chen2019bert,korpusik2019comparison,qin2019stack,zhang2019joint,castellucci2019multi,liu2019cm,pentyala2019multi,krone2020learning,ni2020natural,tang2020end,ren2020intention,wang2020sasgbc,tu2023joint} are rich in this area, there are still areas that could be improved. First, the encoder-based slot-filling classification task is harder since it is not able to fill predefined slots by rephrasing while slot predefining is much easier for decoder LLMs. Moreover, task combination is more versatile for decoder LLMs due to the next token generation architecture. Liu et al.~\cite{liu2023mftcoder} introduced a method for consolidating multiple fine-tuned large language models (LLMs) into a single model by leveraging data balancing, parameter-efficient fine-tuning (PEFT), raw padding, and customized loss functions. Unlike their approach, which targets text generation with varied task inputs, our study focuses on classification tasks with consistent input formats across different tasks. This distinction highlights the unique contributions and specific focus areas of our work.

%% file: sections/3.0_Models.tex
\section{Models}
This study employs a diverse range of models for text classification tasks, including encoder-only models and decoder LLMs, both openly-available and close-source, of varying sizes. The models under investigation are RoBERTa, Llama2, Llama3, GPT-3.5, GPT-4, and GPT-4o.

\subsection{Encoder Models}
RoBERTa-base and RoBERTa-large~\cite{liu2019roberta}, variants of the RoBERTa model, serve as our baseline encoder-only models. These models have demonstrated state-of-the-art performance across various NLP tasks. RoBERTa-base, designed for efficiency and ease of deployment, offers a more compact architecture, while RoBERTa-large, with its increased parameter count, provides enhanced performance on complex tasks. Both models are well-suited for classification tasks, including news categorization, intent detection, and slot-filling classification. Table~\ref{tbl:roberta-models} presents the technical details of RoBERTa-base and RoBERTa-large, aligned with parameters reported in previous research~\cite{liu2019roberta}.

\subsection{Decoder LLMs}
Our study incorporates both openly-available and closed-source decoder LLMs to provide a comprehensive comparison of performance across different model types and sizes. The openly-available decoder LLMs utilized in this research include Llama2-7B, Llama2-70B~\cite{touvron2023llama}, Llama3-8B, and Llama3-70B. These models have achieved remarkable results on various NLP tasks, due to their large-scale pre-training on high-quality datasets. The Llama models, with their transformer architecture, are highly suitable for classification tasks. Our study evaluates the performance of these models on classification tasks both with and without fine-tuning. Table~\ref{tbl:LLM-models} provides the technical details of the Llama models, consistent with previous research~\cite{touvron2023llama}.

We also include OpenAI's family of LLMs in our study: GPT-3.5, GPT-4, and GPT-4o. These closed-source, advanced language models represent the cutting edge of natural language processing. GPT-3.5, an intermediate model building upon GPT-2, offers improved performance across a wide range of NLP tasks while maintaining a smaller footprint. GPT-4 further advances this progress, demonstrating even more impressive results on benchmarks such as language translation, question answering, and text generation. GPT-4o (GPT-4 Omni) is a variant of GPT-4, providing researchers and developers with a powerful tool for exploring and expanding the capabilities of large language models.

All these LLMs are well-suited for classification tasks, and their inclusion in our research allows for a comprehensive comparison between openly-available and close-source LLMs in the context of single- and multi-class text classification. Next, we explained the datasets employed and the experimental procedures conducted in our study, providing insights into the comparative performance of these diverse model types on text classification tasks.

%% file: sections/3.1_Experiments.tex
\section{Experiments}

\subsection{Datasets}
To conduct a comprehensive evaluation of text classification performance across various model types, we selected two openly-available datasets: the 20 Newsgroups (20NG) dataset\footnote{http://qwone.com/~jason/20Newsgroups/}~\cite{LANG1995331} and the MASSIVE en-US dataset\footnote{https://huggingface.co/datasets/AmazonScience/massive}~\cite{fitzgerald2022massive}. These datasets were chosen for their suitability in performing multi-task classification on the same text input, allowing for a thorough comparison of model performance across different classification tasks.

\subsubsection{20 Newsgroups (20NG) Dataset}
We utilized the second version of the 20NG dataset, aligning with previous work~\cite{lin2021bertgcn}. This dataset comprises 18,846 news documents categorized into 20 distinct newsgroups, representing a twenty-class classification task. The documents are sorted by date, with duplicates and certain headers removed to ensure data quality and relevance. To expand our analysis, we derived an additional classification task from the original dataset. By extracting the root labels from the original twenty categories, we created a new seven-class classification task. For instance, the original label 'talk.religion.misc' was reduced to the root label 'talk'. This approach allowed us to evaluate model performance on both twenty-class and seven-class classification tasks. The dataset's original split of 60\% training and 40\% test sets was maintained in our study for both classification tasks.

\subsubsection{MASSIVE en-US Dataset}
The MASSIVE en-US dataset contains 16,521 data points, split into train (11,514 data points), dev (2,033 data points), and test (2,974 data points) sets. For our study, we utilized only the train and test sets, as we did not perform hyperparameter tuning or model selection that would require the dev set.
The text data (utters) is derived base on over 100 hours of audio recordings from more than 1,000 speakers, providing a diverse range of utterances. We employed this text-based dataset for two specific classification tasks: intent classification and slot-filling classification.
In intent classification, each utterance is classified into one of 60 ground truth intent labels.
In slot-filling classification, each word in an utterance is classified into one of 56 ground truth slot labels, with an additional "Other" category for words that do not match any predefined slot. We implemented the BIO (Beginning, Inside, Outside) format for this task, where B indicates the beginning of a slot, I represents a word inside a slot, and O denotes a word outside any slot.

The use of these two datasets allows us to evaluate model performance across a range of classification tasks, from seven-class to twenty-class classification, as well as intent detection and slot-filling, providing a comprehensive benchmark for comparing encoder-only models and LLMs in text classification scenarios. In the following sections, we will detail the prompts used for these datasets and the experimental procedures employed in our study.

\subsection{Prompts}
To direct the LLMs in generating correct labels for the classification tasks, we designed prompts specifically tailored for the 20NG and MASSIVE datasets. Our prompt design strategy focused on creating clear and concise prompts that would direct the LLMs to match the content of each document to the most relevant category from the available options. Take the Llama model fine-tuning on the 20NG dataset as an example, the prompts followed a structured format, with [INST] and [/INST] serving as the pre- and post-query template defined by the creators of the aligned LLMs. Within the prompts, we used variables such as {News} and {Label} to be replaced by the actual document content and ground truth label for each data point. As shown in Listing~\ref{lst:20NG-single-model-prompts} in the Appendix, the prompt instructs the LLM to consider the given news article and assign the most suitable category from the 20 available options. In addition to the zero-shot prompts, we also explored five-shot prompts for all the classification tasks, including the 20NG and MASSIVE datasets, as well as the consolidated model that combined multiple classification tasks. The prompt details for all these variations are provided in the Appendix.

Next, we fed the prompts on the 20NG and MASSIVE datasets to pre-trained instruction-tuned and fine-tuned LLMs asking them to classify the content to the most relevant option.

\subsection{Pre-trained Instruction-tuned LLMs}
To make a comprehensive comparison, we adopted a diverse set of pre-trained instruction-tuned LLMs, ranging from small to large, open to closed-source models. We directly used the prompts mentioned above as the input and fed them into these pre-trained instruction-tuned LLMs and obtain the generated output predictions for news categories classification on the 20NG dataset and for intent and slot-filling classification on the MASSIVE dataset, as well as the multi-task classification on each of the two datasets. After evaluating the performance of these pre-trained LLMs, we then fine-tuned a subset of the models to further assess their capabilities on the text classification tasks, as detailed in the following section.

\subsection{Encoder Model Fine-tuning}
As a baseline, we fine-tuned RoBERTa-base and RoBERTa-large models on both the 20NG and MASSIVE datasets. All tasks were treated as classification problems, including the topic classification in the 20NG dataset, as well as the intent detection and slot-filling tasks in the MASSIVE dataset. For the fine-tuning process, we added a linear classification layer on top of the pre-trained RoBERTa models. In the single-task setting, we used either a sequence classification head (for the 20NG topic classification and MASSIVE intent detection tasks) or a token classification head (for the MASSIVE slot-filling task). In the multi-task setting, we employed either two sequence classification heads (for the joint 20NG topic classifications) or a combination of sequence and token classification heads (for the joint MASSIVE intent detection and slot-filling tasks). These setups are similar to prior work on joint classification models~\cite{chen2019bert}.

For data processing, we leveraged the native RoBERTa tokenizer for non-slot-filling tasks, and the WordPiece RoBERTa tokenizer (RobertaTokenizerFast) to ensure consistency between tokens and labels. During training, we used cross-entropy loss for single-task classification and equally weighted cross-entropy losses for multi-task classification. The training hyperparameters included a batch size of 32, a learning rate of 1e-5 with the Adam optimizer, and 10 epochs for the 20NG dataset and 20 epochs for the MASSIVE dataset.

After fine-tuning, we evaluated the performance of the RoBERTa-base and RoBERTa-large models on the corresponding test sets. We report the topic/intent accuracy and slot-filling F1-score metrics as defined in prior work~\cite{fitzgerald2022massive}. While we briefly explored hyperparameter tuning, the results from the initial set of hyperparameters showed consistency with benchmarks reported in previous studies~\cite{chen2019bert, fitzgerald2022massive}. Therefore, we only report the findings based on the hyperparameter set described above.

\subsection{LLMs Fine-tuning}

In addition to the RoBERTa baselines, we fine-tuned Llama2-7B, Llama2-70B, Llama3-8B, and Llama3-70B, on the 20NG and MASSIVE datasets. For these experiments, we explored both zero-shot and five-shot prompt engineering approaches.

For the twenty-class and seven-class classification tasks on the 20NG dataset, we fine-tuned the decoder LLMs using a full fine-tuning approach, as prior studies have shown this to generally outperform LoRA in terms of accuracy and sample efficiency~\cite{lv2023full, pan2024lisa, balazy2024lora}. The fine-tuning process involved using the same prompts as in the earlier experiments with the pre-trained, instruction-tuned LLMs. During training, the full prompt, including the pre- and post-query template ([INST] and [/INST]) and the text content, was fed to the LLMs. The models then generated tokens, which were compared to the ground truth label after the post-query template to compute the cross-entropy loss. The LLMs then updated their weights based on this loss to adapt to the classification tasks. For the evaluation on the test set, we used a similar prompt structure, but without the ground truth label after the post-query template. Instead, we assessed the performance of the fine-tuned LLMs based on the generated tokens and the actual ground truth label.

The fine-tuning procedure for the MASSIVE dataset followed a similar approach. For the intent classification task, the process was the same as the 20NG classification tasks. However, for the slot-filling task, the ground truth labels were not single strings, but rather a list of strings indicating the slot labels for each word in the utterance. The training and testing processes were adapted accordingly to handle this structured output.

\subsection{Model Consolidation}

To explore the multi-task capabilities of the LLMs, we consolidated the twenty-class and seven-class topic classification tasks from the 20NG dataset into a single unified model and similarly consolidated the intent detection and slot-filling tasks from the MASSIVE dataset into a single consolidated model. We designed a unified prompt with two JSON output fields. This approach allowed the LLMs to perform both classification tasks simultaneously, rather than treating them as separate models. The consolidated prompt instructed the LLMs to generate a JSON string with two key-value pairs: \{"task1": "task1\_label", "task2": "task2\_label"\}. Take the 20NG dataset as an example. Given a news document, the LLMs were asked to classify it using both the twenty-class and seven-class classification tasks, with the generated output providing the predicted labels for each task. The LLM was asked to replace "task1\_label" with one of the labels in the twenty-class classification and replace "task2\_label" with one of the labels in the seven-class classification. As shown in Listing~\ref{lst:consolidated-nodel-prompts-five-shot} in the Appendix, the five-shot prompts for this multi-task setting included an additional variable $\{$Examples$\}$, which represented five examples of news documents and their ground truth label pairs from the training set. This was intended to provide the LLMs with relevant context to guide their joint classification predictions.

\subsection{Fine-tuned LLMs Implementation Details}
For the baseline models, we implemented RoBERTa-base and RoBERTa-large using the parameter settings from prior research~\cite{liu2019roberta}. The detailed configuration of these RoBERTa models is provided in Table~\ref{tbl:roberta-models}.

\begin{table}[ht]
\centering
\def\arraystretch{1.4}

\scalebox{0.8}{
\begin{tabular}{|c|c|c|}
\hline
  Parameters     & RoBERTa-base   & RoBERTa-large \\ \hline
hidden$\_$size & 768 & 1024    \\ \hline
num$\_$attention\_heads & 12 & 16    \\ \hline
intermediate$\_$size & 12 &  24     \\ \hline
max$\_$position$\_$embeddings & 3072 & 4096    \\ \hline
type$\_$vocab$\_$size & \multicolumn{2}{|c|}{514}    \\ \hline
num$\_$hidden$\_$layers & \multicolumn{2}{|c|}{1}     \\ \hline
vocab$\_$size & \multicolumn{2}{|c|}{50265}    \\ \hline
hidden$\_$dropout$\_$prob & \multicolumn{2}{|c|}{0.1}   \\ \hline
attention$\_$probs$\_$dropout$\_$prob & \multicolumn{2}{|c|}{0.1}     \\ \hline
hidden$\_$act & \multicolumn{2}{|c|}{GELU}    \\ \hline
initializer$\_$range & \multicolumn{2}{|c|}{0.02}    \\ \hline
layer$\_$norm$\_$eps & \multicolumn{2}{|c|}{1e-5}     \\ \hline
\end{tabular}
}
\caption{Parameters for RoBERTa-base and RoBERTa-large~\cite{liu2019roberta}.}
\label{tbl:roberta-models}
\end{table}


To fine-tune the decoder LLMs, we utilized the openly-available Llama2 and Llama3 model checkpoints. The fine-tuning was performed on 8 NVIDIA A100 80GB GPUs. The fine-tuning duration ranged from hours to 3 days, depending on the model size and training data volume. All models were trained until convergence. The fine-tuning hyperparameters were adjusted based on the model size. We referred to Llama2-7B and Llama3-8B as the "small" models, and Llama2-70B and Llama3-70B as the "large" models. The specific hyperparameter settings were as follows:
\begin{itemize}
\item Small models: 4,000 training steps, 0 warm-up, 1e-6 learning rate, 4,096 model dimension, 32 attention heads, 8 key-value heads, 32 layers, 16,384 maximum input length, and a batch size of 8.
\item Large models: 8,000 training steps, 0 warm-up, 1e-6 learning rate, 8,192 model dimension, 64 attention heads, 8 key-value heads, 80 layers, 16,384 maximum input length, and a batch size of 8.
\end{itemize}
Both small and large models used SwiGLU as the nonlinear activation function and AdamW as the optimizer. Additionally, row padding was applied to improve training efficiency~\cite{liu2023mftcoder}. The fine-tuning implementation for both the RoBERTa and Llama models was done using the PyTorch deep learning API. The detailed model configurations are summarized in Table~\ref{tbl:LLM-models}.

\begin{table}[ht]
\centering
\def\arraystretch{1.4}
\label{tab:llama_params}
\scalebox{0.68}{
\begin{tabular}{|c|c|c|c|c|}
\hline
& Llama2-7B & Llama2-70B & Llama3-8B & Llama3-70B \\
\hline
Layers & 32 & 80 & 32 & 80  \\ \hline
Model Dimension & 4096 & 8192 & 4096 & 8192 \\
\hline
FFN Dimension & 6144 & 12288 & 6144 & 12288 \\
\hline
Attention Heads & 32 & 64 & 32 & 64 \\
\hline

Peak Learning Rate & \multicolumn{4}{|c|}{$1 \times 10^{-6}$ } \\
\hline
Key/Value Heads & \multicolumn{4}{|c|}{8} \\
\hline
Activation Function & \multicolumn{4}{|c|}{SwiGLU} \\
\hline
Vocabulary Size & \multicolumn{4}{|c|}{128,000} \\
\hline
Positional Embeddings & \multicolumn{4}{|c|}{RoPE ($\theta = 500,000$)} \\
        
\hline
\end{tabular}
}
\caption{Comparison of LLaMA Model Parameters. We display the configuration for LLaMA2-7B, LLaMA2-70B, LLaMA3-8B, and LLaMA3-70B models.}
\label{tbl:LLM-models}

\end{table}

\begin{table*}[ht]
\centering
\def\arraystretch{1.5}
\scalebox{0.78}{
\begin{tabular}{|c|cc|cc|cc|cc|}
\hline
\multirow{2}{*}{Model} & \multicolumn{2}{c|}{20 NG Dual Model} & \multicolumn{2}{c|}{20 NG Consolidated Model} & \multicolumn{2}{c|}{MASSIVE Dual Model} & \multicolumn{2}{c|}{MASSIVE Consolidated Model}\\ 
\cline{2-9}
     & \multicolumn{1}{c|}{20-class} & 7-class & \multicolumn{1}{c|}{20-class} & 7-class & \multicolumn{1}{c|}{Intent} & Slot Filling & \multicolumn{1}{c|}{Intent} & Slot Filling\\ 
\hline
RoBERTa-base  & \multicolumn{1}{c|}{86.7\%} & 93.6\% & \multicolumn{1}{c|}{86.5\%} & 93.6\% & \multicolumn{1}{c|}{88.9\%} & 83.4\% & \multicolumn{1}{c|}{89.1\%} & 83.6\% \\ \hline
RoBERTa-large & \multicolumn{1}{c|}{88.1\%} & 94.2\% & \multicolumn{1}{c|}{87.8\%} & 94.6\% & \multicolumn{1}{c|}{90.1\%} & 85.0\% & \multicolumn{1}{c|}{90.0\%} & 84.7\% \\ \hline
Llama2-7B Five-shot w/o FT         & \multicolumn{1}{c|}{0\%} & 0\% & \multicolumn{1}{c|}{0\%} & 0\% & \multicolumn{1}{c|}{35.8\%} & 4.6\% & \multicolumn{1}{c|}{7.2\%} & 7.9\%\\ \hline
Llama2-7B Zero-shot w/o FT         & \multicolumn{1}{c|}{25.1\%} & 23.8\% & \multicolumn{1}{c|}{11.4\%} & 21.8\% & \multicolumn{1}{c|}{30.3\%} & 0\% & \multicolumn{1}{c|}{0\%} & 0\%\\ \hline
Llama2-70B Five-shot w/o FT       & \multicolumn{1}{c|}{0\%} & 0\% & \multicolumn{1}{c|}{0\%} & 0\% & \multicolumn{1}{c|}{70.8\%} & 9.5\% & \multicolumn{1}{c|}{62.7\%} & 5.1\% \\ \hline
Llama2-70B Zero-shot w/o FT        & \multicolumn{1}{c|}{64.1\%} & 44.9\% & \multicolumn{1}{c|}{48.6\%} & 62.7\% & \multicolumn{1}{c|}{60.8\%} & 0\% & \multicolumn{1}{c|}{0\%} & 0\% \\ \hline
Llama2-7B Five-shot w/ FT  & \multicolumn{1}{c|}{88.2\%} & 94.6\% & \multicolumn{1}{c|}{88.1\%} & 93.9\% & \multicolumn{1}{c|}{90.3\%} & 84.0\% & \multicolumn{1}{c|}{89.8\%} & 84.1\%\\ \hline
Llama2-7B Zero-shot w/ FT  & \multicolumn{1}{c|}{89.5\%} & 95.2\% & \multicolumn{1}{c|}{89.2\%} & 94.9\% & \multicolumn{1}{c|}{90.4\%} & 82.5\% & \multicolumn{1}{c|}{89.6\%} & 83.4\% \\ \hline
Llama2-70B Five-shot w/ FT & \multicolumn{1}{c|}{88.4\%} & 95.3\% & \multicolumn{1}{c|}{88.7\%} & 94.6\% & \multicolumn{1}{c|}{90.0\%} & 83.9\% & \multicolumn{1}{c|}{90.0\%} & 84.3\% \\ \hline
Llama2-70B Zero-shot w/ FT & \multicolumn{1}{c|}{91.2\%} & 96.3\% & \multicolumn{1}{c|}{91.5\%} & 96.6\% & \multicolumn{1}{c|}{90.2\%} & 83.1\% & \multicolumn{1}{c|}{90.4\%} & 83.6\% \\ \hline
Llama3-8B Five-shot w/o FT        & \multicolumn{1}{c|}{0\%} & 0\% & \multicolumn{1}{c|}{0\%} & 0\% & \multicolumn{1}{c|}{69.7\%} & 2.0\% & \multicolumn{1}{c|}{38.9\%} & 5.9\% \\ \hline
Llama3-8B Zero-shot w/o FT        & \multicolumn{1}{c|}{52.2\%} & 41.2\% & \multicolumn{1}{c|}{34.1\%} & 51.6\% & \multicolumn{1}{c|}{61.7\%} & 1.6\% & \multicolumn{1}{c|}{0\%} & 0\% \\ \hline
Llama3-70B Five-shot w/o FT       & \multicolumn{1}{c|}{0\%} & 0\% & \multicolumn{1}{c|}{0\%} & 0\% & \multicolumn{1}{c|}{80.6\%} & 20.6\% & \multicolumn{1}{c|}{70.7\%} & 16.4\% \\ \hline
Llama3-70B Zero-shot w/o FT       & \multicolumn{1}{c|}{69.5\%} & 67\% & \multicolumn{1}{c|}{56.7\%} & 78\% & \multicolumn{1}{c|}{78.9\%} & 12.2\% & \multicolumn{1}{c|}{0\%} & 0\% \\ \hline
Llama3-8B Five-shot w/ FT  & \multicolumn{1}{c|}{91.1\%} & 95.9\% & \multicolumn{1}{c|}{91\%} & 96.1\% & \multicolumn{1}{c|}{90.6\%} & 86.2\% & \multicolumn{1}{c|}{89.7\%} & 84.5\% \\ \hline
Llama3-8B Zero-shot w/ FT  & \multicolumn{1}{c|}{91.6\%} & 96.3\% & \multicolumn{1}{c|}{90.8\%} & 96.2\% & \multicolumn{1}{c|}{90.6\%} & 85.6\% & \multicolumn{1}{c|}{90.0\%} & 85.5\% \\ \hline
\textbf{Llama3-70B Five-shot w/ FT} & \multicolumn{1}{c|}{\textbf{91.9\%}} & \textbf{96.5\%} & \multicolumn{1}{c|}{\textbf{92.1\%}} & \textbf{96.8\%} & \multicolumn{1}{c|}{\textbf{90.8\%}} & \textbf{86.0\%} & \multicolumn{1}{c|}{\textbf{90.2\%}} & \textbf{86.8\%} \\ \hline
Llama3-70B Zero-shot w/ FT & \multicolumn{1}{c|}{91.5\%} & 96.1\% & \multicolumn{1}{c|}{91.2\%} & 96.2\% & \multicolumn{1}{c|}{90.8\%} & 85.8\% & \multicolumn{1}{c|}{89.8\%} & 86.3\% \\ \hline
GPT3.5 Five-shot w/o FT   & \multicolumn{1}{c|}{0\%} & 0\% & \multicolumn{1}{c|}{0\%} & 0\% & \multicolumn{1}{c|}{64.0\%} & 2.0\% & \multicolumn{1}{c|}{73.6\%} & 21.0\% \\ \hline
GPT3.5 Zero-shot w/o FT   & \multicolumn{1}{c|}{72.9\%} & 55.1\% & \multicolumn{1}{c|}{62.3\%} & 69.6\% & \multicolumn{1}{c|}{60.7\%} & 1.8\% & \multicolumn{1}{c|}{69.6\%} & 14.2\% \\ \hline
GPT4 Five-shot w/o FT   & \multicolumn{1}{c|}{0\%} & 0\% & \multicolumn{1}{c|}{0\%} & 0\% & \multicolumn{1}{c|}{83.3\%} & 37.1\% & \multicolumn{1}{c|}{79.9\%} & 28.1\% \\ \hline
GPT4 Zero-shot w/o FT   & \multicolumn{1}{c|}{80.4\%} & 74.5\% & \multicolumn{1}{c|}{80.6\%} & 88.5\% & \multicolumn{1}{c|}{81.9\%} & 13.6\% & \multicolumn{1}{c|}{76.9\%} & 21.6\% \\ \hline
GPT4o Five-shot w/o FT  & \multicolumn{1}{c|}{0\%} & 0\% & \multicolumn{1}{c|}{0\%} & 0\% & \multicolumn{1}{c|}{82.0\%} & 18.9\% & \multicolumn{1}{c|}{81.0\%} & 39.4\% \\ \hline
GPT4o Zero-shot w/o FT  & \multicolumn{1}{c|}{77.4\%} & 73.3\% & \multicolumn{1}{c|}{73.6\%} & 84.1\% & \multicolumn{1}{c|}{62.7\%} & 14.8\% & \multicolumn{1}{c|}{0\%} & 0\% \\ \hline
\end{tabular}
}
\caption{Comparison of encoder models and decoder LLMs with and without fine-tuning (FT) on twenty-class and seven-class classification tasks of the 20NG dataset, and intent and slot-filling classification tasks on the MASSIVE dataset. We evaluated the performance of the 20-class and 7-class classifications on the 20NG dataset, as well as intent classification on the MASSIVE dataset, using accuracy as the metric. For the slot-filling task on the MASSIVE dataset, we used the F1-score to measure performance. The Llama3-70B five-shot fine-tuned model (in bold) has achieved the best performance on both datasets.}
\label{tbl:news-results}
\end{table*}

\subsection{Evaluation Results}
To align the evaluation metrics with prior work~\cite{lin2021bertgcn, fitzgerald2022massive}, we adopted accuracy to measure the performance on the newsgroup classification task in the 20NG dataset and the intent classification task in the MASSIVE en-US dataset. For the slot-filling classification in the MASSIVE en-US dataset, we used the F1-score metric.
The F1-score calculation for the slot-filling task was performed using the f1\_score function from the seqeval.metrics library. This approach groups consecutive slot labels into entities, ignores 'Other' labels, appends padding labels, and truncates extra labels before the evaluation. This ensures an accurate measurement of the slot-filling performance.

The evaluation results are presented in Table~\ref{tbl:news-results}. On the 20NG dataset, the fine-tuned Llama3-70B model with five-shot prompts achieved an accuracy of 91.9\% on the twenty-class task and 96.5\% on the seven-class task, outperforming all other models, including the RoBERTa-large baselines. These results suggest that the larger LLMs (e.g., Llama3-70B) generally performed better than their smaller counterparts (e.g., Llama3-8B), and the Llama3 models outperformed the Llama2 models. Interestingly, the performance difference between the fine-tuned large LLMs (e.g., Llama3-70B) and small LLMs (e.g., Llama3-8B) was not substantial. Additionally, the comparison of openly-available and closed-source pre-trained, instruction-tuned LLMs showed that GPT-4 and GPT-4o achieved better performance than the Llama models.
When we merged the twenty-class and seven-class tasks into a single consolidated model, it achieved on-par performance (92.1\% accuracy on the twenty-class task and 96.8\% accuracy on the seven-class task) compared to the dual-model setup.

The evaluation on the MASSIVE en-US dataset yielded similar results. The fine-tuned Llama3-70B model with five-shot prompts performed the best, achieving a 90.8\% accuracy on the intent classification task and an 86.0\% F1-score on the slot-filling task. The consolidated multi-task model also matched the performance of the dual-model setup, with an accuracy of 90.2\% on the intent classification and an F1-score of 86.8\% on the slot-filling task. Overall, the results demonstrate the superior performance of fine-tuned, large LLMs, such as Llama3-70B, on text classification tasks compared to encoder-only models and all other LLMs. Additionally, the consolidated multi-task models were able to achieve equivalent performance to dual-model setups, offering potential benefits in terms of reduced latency and resource utilization.

%% file: sections/4.0_general_discussion.tex
\section{Discussion}
\subsection{LLMs in Classification Task}
It is crucial to understand the distinction between encoder models, like BERT, and decoder LLMs in the context of text classification tasks. Prior research has shown that BERT-like encoder models generally outperform LSTM and traditional machine learning methods on classification tasks~\cite{devlin2018bert, liu2019roberta}. The performance of BERT-like encoder models also surpasses decoder LLMs without fine-tuning~\cite{bucher2024fine}. However, our results demonstrate that fully fine-tuned decoder LLMs, such as Llama3-70B, are able to outperform encoder models like RoBERTa-large on text classification tasks (Table~\ref{tbl:news-results}). This is likely due to the larger parameter size of the fine-tuned decoder LLMs (billion-level) compared to the encoder models (million-level). Larger models with more parameters are generally expected to achieve better performance, especially when coupled with the larger training data (trillion-level tokens) used for the decoder LLMs, compared to the encoder models (billion-level tokens).

Within the Llama family of LLMs, we observed that the Llama3 models consistently outperformed the Llama2 models of the same scale, both in fine-tuned and pre-trained, instruction-tuned settings. This can be attributed to the improved quantity (15T tokens vs. 1.8T tokens) and quality of the training data used for Llama3~\cite{MetaAI2024}. Additionally, our results show that the larger Llama models (70B) generally performed better than their smaller counterparts (8B), for fine-tuned and pre-trained instruction-tuned LLMs. This aligns with previous studies ~\cite{touvron2023llama, MetaAI2024} and suggests that the increased size and complexity of the larger models enable them to learn more intricate patterns in the data, leading to better performance on classification tasks.

Table~\ref{tbl:news-results} also showed that the close-source LLMs such as GPT4 and GPT4o have achieved the best performance among the pre-trained, instruction-tuned models, outperforming the openly-available Llama2 and Llama3 models. While the closed-source models may have an advantage in certain tasks, the openly-available LLMs offer transparency, flexibility, and community-driven development, which can lead to rapid innovation and improvement over time. The performance gap between openly-available and closed-source LLMs may also be task-dependent.

Finally, our results indicate that fine-tuned LLMs exhibit minimal performance differences between zero-shot and five-shot learning (Table~\ref{tbl:news-results}). This suggests that these models are highly adaptable and can generalize well from their pre-training, potentially enabling more efficient and effective training methods for natural language processing and other applications.

\subsection{Application}
The promising performance of the fine-tuned single- and multi-task LLMs on classification tasks has great potential to benefit AI agent applications. First, the superior accuracy of these fine-tuned LLMs compared to state-of-the-art encoder models like RoBERTa and pre-trained instruction-tuned LLMs can enhance the tool selection process, a critical step in AI agent functionality. For instance, an AI agent designed for technical support can more accurately identify the appropriate tool or sub-agent based on the user's query, ensuring a more efficient resolution process. Second, the multi-tasking capabilities of the fine-tuned decoder LLMs can reduce the number of model calls and improve overall efficiency. By integrating tasks such as tool selection and slot-filling, an AI agent can simultaneously determine the best tool and gather relevant user information from the input text, significantly reducing latency. This model consolidation not only speeds up the decision-making process but also enhances the user experience by providing quicker and more accurate responses.

\subsection{Limitations and Future Work}
The present research represents an initial step towards classifying single- and multi-task text input using LLMs. While the results are based on openly-available datasets with consistent evaluation metrics, real-world applications of LLMs often involve a wider range of tasks, such as question-answering, summarization, and translation, in addition to classification. To extend the generalizability of our findings, future research should consider adjusting the datasets and task selection to include a more diverse set of applications. For example, the study could be expanded beyond classification tasks to include generation tasks, such as joint classification and generation tasks (e.g., intent detection and query rewrite). Recognizing and adapting to these variations in task types and datasets is crucial to fully understand the potential and limitations of LLMs in practical AI agent applications. Building upon the insights uncovered from this research, future studies should explore fine-tuning techniques for single- and multi-task LLMs that can account for a wider range of usage scenarios, ultimately improving the accuracy of classification tasks and the quality of text generation.

%% file: sections/5.0_conclusion.tex
\section{Conclusion}
In this comprehensive study, we thoroughly explored the application of LLMs for text classification tasks. We evaluated a diverse range of openly-available and closed-source decoder LLMs, including both small and large models, with and without full parameter fine-tuning. The models were assessed on the openly-available 20NG and MASSIVE en-US classification datasets. The results consistently demonstrated that the fully fine-tuned Llama3-70B model outperformed encoder-only models like RoBERTa-large, as well as other decoder LLMs. Interestingly, we found that the performance difference between fully fine-tuned decoder LLMs of various sizes and versions was relatively minimal. Furthermore, the advantage of few-shot prompt engineering for these fully fine-tuned LLMs was limited. By consolidating the fully fine-tuned LLMs on each dataset into a single multi-task model, we were able to match the performance of dual-model setups in both classification tasks. This consolidated approach offers potential benefits in terms of reduced latency and resource utilization compared to maintaining separate models for each task. Overall, our research advances the understanding of decoder LLMs and their performance on text classification tasks, both with and without fine-tuning. The study provides a comprehensive benchmark of LLM capabilities and introduces a method to consolidate multiple fine-tuned decoder LLMs into a single efficient model. These findings inspire the more widespread use of LLMs in text classification applications within the AI ecosystem.

%% file: sections/6.0_appendix.tex
\section{Appendix}
\begin{lstlisting}[caption={An example of a training zero-shot prompt for 20NG twenty-class classification task.}, label={lst:20NG-single-model-prompts}]
MODEL_TRAIN_INST_PROMPT = """
[INST]You will be given a news text.
Your task is to classify it as one of the labels in the list ['alt.atheism', 'comp.graphics', 'comp.os.ms-windows.misc', 'comp.sys.ibm.pc.hardware', 'comp.sys.mac.hardware', 'comp.windows.x', 'misc.forsale', 'rec.autos', 'rec.motorcycles', 'rec.sport.baseball', 'rec.sport.hockey', 'sci.crypt', 'sci.electronics', 'sci.med', 'sci.space', 'soc.religion.christian', 'talk.politics.guns', 'talk.politics.mideast', 'talk.politics.misc', 'talk.religion.misc']

Output the label only and absolutely nothing else.
Now it is your turn.

Beginning of <<News>>
{News}
End of <<News>>

Label: [/INST]
{Label}"""
\end{lstlisting}

\begin{lstlisting}[caption={An example of a training zero-shot prompt for 20NG seven-class classification task.}, label={lst:single-model-prompts}]
MODEL_TRAIN_INST_PROMPT = """
You will be given a news text.
Your task is to classify it as one of the labels in the list ["alt", "comp", "misc", "rec", "sci", "soc", talk"]

Output the label only and absolutely nothing else.
Now it is your turn.

Beginning of <<News>>
{News}
End of <<News>>

Label:"""
\end{lstlisting}

\begin{lstlisting}[caption={An example of a training five-shot prompt for 20NG twenty-class classification task.}, label={lst:single-model-prompts}]
MODEL_TRAIN_INST_PROMPT = """
[INST]You will be given a news text.
Your task is to classify it as one of the labels in the list ['alt.atheism', 'comp.graphics', 'comp.os.ms-windows.misc', 'comp.sys.ibm.pc.hardware', 'comp.sys.mac.hardware', 'comp.windows.x', 'misc.forsale', 'rec.autos', 'rec.motorcycles', 'rec.sport.baseball', 'rec.sport.hockey', 'sci.crypt', 'sci.electronics', 'sci.med', 'sci.space', 'soc.religion.christian', 'talk.politics.guns', 'talk.politics.mideast', 'talk.politics.misc', 'talk.religion.misc']

Bellow are some examples with the correct classification:

Beginning of <<Examples>>
{{Examples}}
End of <<Examples>>

Output the label only and absolutely nothing else.
Now it is your turn.

Beginning of <<News>>
{News}
End of <<News>>

Label: [/INST]
{Label}"""
\end{lstlisting}

\begin{lstlisting}[caption={An example of a training five-shot prompt for 20NG seven-class classification task.}, label={lst:single-model-prompts}]
MODEL_TRAIN_INST_PROMPT = """
You will be given a news text.
Your task is to classify it as one of the labels in the list ["alt", "comp", "misc", "rec", "sci", "soc", talk"]

Bellow are some examples with the correct classification:

Beginning of <<Examples>>
{{Examples}}
End of <<Examples>>

Output the label only and absolutely nothing else.
Now it is your turn.

Beginning of <<News>>
{News}
End of <<News>>

Label:"""
\end{lstlisting}

\begin{lstlisting}[caption={An example of training zero-shot prompts for the multi-task classification task on 20NG dataset.}, label={lst:consolidated-model-prompts}]
MODEL_TRAIN_INST_PROMPT = """
[INST]You will be given a news text.
You are given two classification tasks.
You should output the result as a two json fields as {"task1": "task1_label", "task2": "task2_label"}
In the first task, given the news text, you are asked to classify it as one of the labels in the list ["alt", "comp", "misc", "rec", "sci", "soc", talk"] and change task1_label to the correct label in the list.
In the second task, given the news text, you are asked to classify it as one of the labels in the list ['alt.atheism', 'comp.graphics', 'comp.os.ms-windows.misc', 'comp.sys.ibm.pc.hardware', 'comp.sys.mac.hardware', 'comp.windows.x', 'misc.forsale', 'rec.autos', 'rec.motorcycles', 'rec.sport.baseball', 'rec.sport.hockey', 'sci.crypt', 'sci.electronics', 'sci.med', 'sci.space', 'soc.religion.christian', 'talk.politics.guns', 'talk.politics.mideast', 'talk.politics.misc', 'talk.religion.misc'] and change task2_label to the correct label in the list.

Output the two json fields only and absolutely nothing else.
Now it is your turn.

Beginning of <<News>>
{News}
End of <<News>>

Output: [/INST]
{Label}"""
\end{lstlisting}

\begin{lstlisting}[caption={An example of training five-shot prompts for the multi-task classification task on 20NG dataset.}, label={lst:consolidated-nodel-prompts-five-shot}]
MODEL_TRAIN_INST_PROMPT = """
[INST]You will be given a news text.
You are given two classification tasks.
You should output the result as a two json fields as {"task1": "task1_label", "task2": "task2_label"}
In the first task, given the news text, you are asked to classify it as one of the labels in the list ["alt", "comp", "misc", "rec", "sci", "soc", talk"] and change task1_label to the correct label in the list.
In the second task, given the news text, you are asked to classify it as one of the labels in the list ['alt.atheism', 'comp.graphics', 'comp.os.ms-windows.misc', 'comp.sys.ibm.pc.hardware', 'comp.sys.mac.hardware', 'comp.windows.x', 'misc.forsale', 'rec.autos', 'rec.motorcycles', 'rec.sport.baseball', 'rec.sport.hockey', 'sci.crypt', 'sci.electronics', 'sci.med', 'sci.space', 'soc.religion.christian', 'talk.politics.guns', 'talk.politics.mideast', 'talk.politics.misc', 'talk.religion.misc'] and change task2_label to the correct label in the list.

Bellow are some examples with the correct classification:

Beginning of <<Examples>>
{Examples}
End of <<Examples>>

Output the two json fields only and absolutely nothing else.
Now it is your turn.

Beginning of <<News>>
{News}
End of <<News>>

Output: [/INST]
{Label}"""
\end{lstlisting}

\begin{lstlisting}[caption={An example of a training zero-shot prompt for MASSIVE intent classification task.}, label={lst:single-model-prompts}]
MODEL_TRAIN_ZERO_SHOT_INST_PROMPT = """
You will be given an utter text.
Your task is to classifity it as one of the labels in the list {Labels}.

Output the label only and absolutely nothing else.
Now it is your turn.

Beginning of <<Utter>>
{Utter}
End of <<Utter>>

Label:"""
\end{lstlisting}

\begin{lstlisting}[caption={An example of a training five-shot prompt for MASSIVE intent classification task.}, label={lst:single-model-prompts}]
MODEL_TRAIN_FIVE_SHOT_INST_PROMPT = """
You will be given an utter text.
Your task is to classifity it as one of the labels in the list {Labels}.

Bellow are some examples with the correct classification:

Beginning of <<Examples>>
{Examples}
End of <<Examples>>

Output the label only and absolutely nothing else.
Now it is your turn.

Beginning of <<Utter>>
{Utter}
End of <<Utter>>

Label:"""
\end{lstlisting}

\begin{lstlisting}[caption={An example of a training zero-shot prompt for MASSIVE slot-filling classification task.}, label={lst:single-model-prompts}]
MODEL_TRAIN_ZERO_SHOT_INST_PROMPT = """
You will be given an utter text.
Your task is to classify each of the word in the utter text as one of the labels in the list {Labels}.
If none of the labels in the list are correct, then output the label for that word as "Other".
Put all the labels in the slots list.

Output the Slots only and absolutely nothing else.
Now it is your turn.

Beginning of <<Utter>>
{Utter}
End of <<Utter>>

Slots:"""
MODEL_TRAIN_ZERO_SHOT_INST_PROMPT = """
You will be given an utter text.
Your task is to classify each of the word in the utter text as one of the labels in the list {Labels}.
If none of the labels in the list are correct, then output the label for that word as "Other".
Put all the labels in the slots list.

Output the Slots only and absolutely nothing else.
Now it is your turn.

Beginning of <<Utter>>
{Utter}
End of <<Utter>>

Slots:"""
\end{lstlisting}

\begin{lstlisting}[caption={An example of a training five-shot prompt for MASSIVE slot-filling classification task.}, label={lst:single-model-prompts}]
MODEL_TRAIN_FIVE_SHOT_INST_PROMPT = """
You will be given an utter text.
Your task is to classify each of the word in the utter text as one of the labels in the list {Labels}.
If none of the labels in the list are correct, then output the label for that word as "Other".
Put all the labels in the slots list.

Bellow are some examples with the correct slot filling classification:

Beginning of <<Examples>>
{Examples}
End of <<Examples>>

Output the Slots only and absolutely nothing else.
Now it is your turn.

Beginning of <<Utter>>
{Utter}
End of <<Utter>>

Slots:"""
\end{lstlisting}

\begin{lstlisting}[caption={An example of a training zero-shot prompt for MASSIVE multi-task classification.}, label={lst:single-model-prompts}]
MODEL_TRAIN_ZERO_SHOT_INST_SYSTEM_PROMPT = """
You will be given an utter text.
You are given two classification tasks.
You should output the result as a two json fields as {"task1": "task1_label", "task2": "task2_label"}
"""

MODEL_TRAIN_ZERO_SHOT_INST_QUERY_PROMPT = """
In the first task, given the utter text, you are asked to classify it as one of the labels in the list {Intent_Labels}.
In the second task, given the utter text, you are asked to classify each of the word in the utter text as one of the labels in the list {Slot_Filling_Labels}. If none of the labels in the list are correct, then output the label for that word as "Other". Put all the labels in the slots list.

Output the two json fields only and absolutely nothing else.
Now it is your turn.

Beginning of <<Utter>>
{Utter}
End of <<Utter>>

Output:"""
\end{lstlisting}

\begin{lstlisting}[caption={An example of a training five-shot prompt for MASSIVE multi-task classification.}, label={lst:single-model-prompts}]
MODEL_TRAIN_FIVE_SHOT_INST_SYSTEM_PROMPT = """
You will be given an utter text.
You are given two classification tasks.
You should output the result as a two json fields as {"task1": "task1_label", "task2": "task2_label"}
"""

MODEL_TRAIN_FIVE_SHOT_INST_QUERY_PROMPT = """
In the first task, given the utter text, you are asked to classify it as one of the labels in the list {Intent_Labels}.
In the second task, given the utter text, you are asked to classify each of the word in the utter text as one of the labels in the list {Slot_Filling_Labels}. If none of the labels in the list are correct, then output the label for that word as "Other". Put all the labels in the slots list.

Bellow are some examples with the correct classification:

Beginning of <<Examples>>
{Examples}
End of <<Examples>>

Output the two json fields only and absolutely nothing else.
Now it is your turn.

Beginning of <<Utter>>
{Utter}
End of <<Utter>>

Output:"""
\end{lstlisting}